\documentclass{article}
\usepackage[preprint]{neurips_2026}
\usepackage[utf8]{inputenc}
\usepackage[T1]{fontenc}
\usepackage{hyperref}
\usepackage{url}
\usepackage{booktabs}
\usepackage{amsfonts}
\usepackage{nicefrac}
\usepackage{microtype}
\usepackage{xcolor}
\usepackage{graphicx}
\usepackage{amsmath}
\usepackage{amssymb}
\usepackage{multirow}
\usepackage{array}
\usepackage{pifont}

\definecolor{neuripsblue}{rgb}{0.21,0.49,0.74}
\hypersetup{breaklinks,colorlinks,allcolors=neuripsblue}

\title{Do Vision Models Learn Physical Constraints or Rendering Shortcuts? A Counterfactual Benchmark for Grounded Physical Consistency}
\author{\textbf{M. Moein Esfahani$^{1}$} \quad \textbf{Sepehr Salem$^{2}$} \\
\textbf{Mohammed Alser$^{2}$} \quad \textbf{Vince Calhoun$^{3}$} \\
\small $^{1,3}$ Tri-institutional Center for Translational Research in Neuroimaging and Data Science (TReNDS) \\
\small Georgia State University, Georgia Institute of Technology, and Emory University, Atlanta, GA, USA \\
\small $^{2}$ Georgia State University, Atlanta, GA, USA \\
\small \texttt{mesfahani1@gsu.edu}; \texttt{ssalemghahfarokhi1@gsu.edu} \\
\small \texttt{vcalhoun@gsu.edu}}

\begin{document}
\maketitle

\begin{abstract}
Modern image editing models can satisfy a text instruction while breaking the
physics of the edited scene. A new object may cast no shadow, a mirror may
fail to reflect visible geometry, or an object may float above a surface that
should support it. We study physical plausibility diagnosis, detecting
whether an edited image violates scene physics, naming the violation type,
localizing the affected region, and explaining the failure in language. We
introduce a counterfactual benchmark whose controlled synthetic component uses
Mitsuba~3 to generate 5,500 images from 500 scene families. Each family
contains one clean image and ten matched violations involving shadows,
reflection, support, surface response, and occlusion. The renderer pipeline provides
category labels, affected-region masks and boxes, scene metadata, and
explanation targets. We use LLaVA-1.5-7B, Qwen2.5-VL-7B, and InternVL3.5-8B as
diagnostic baselines rather than proposed methods. On a 1,650-image synthetic
test set, the adapted baselines reach 64.0--67.8\% category macro-F1 on
standard held-out scenes. For LLaVA-1.5-7B, category macro-F1 falls from
64.0\% on the standard split to 40.8\% under intervention shift. This gap shows
that high in-distribution accuracy partly reflects cues tied to rendering and
counterfactual construction.
\end{abstract}

\begin{figure*}[!t]
\centering
\includegraphics[width=\textwidth]{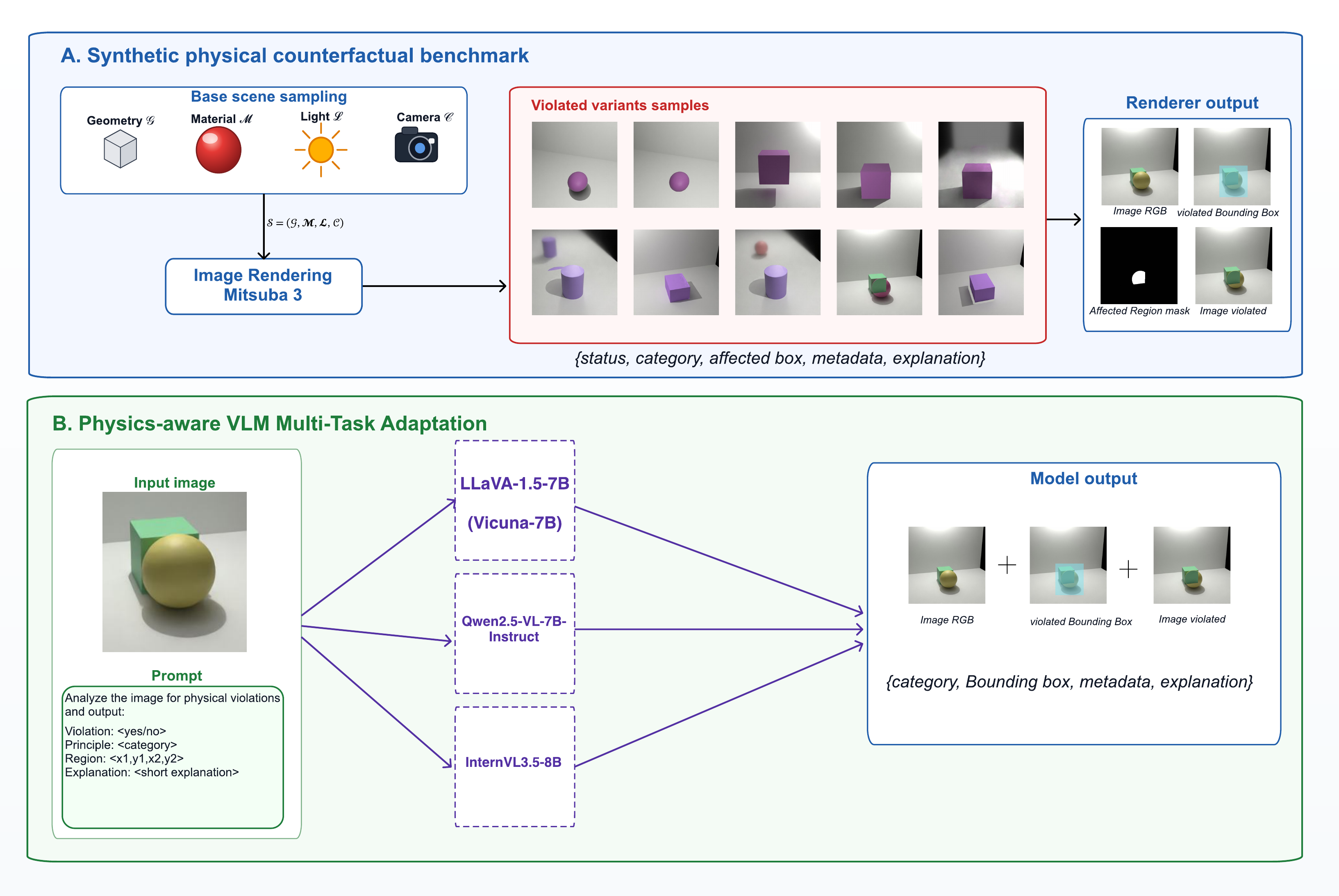}
\caption{Overview of the benchmark and diagnostic baselines. (A) Controlled rendering creates
clean scenes, physical counterfactuals, and category, region, and explanation
supervision. (B) LLaVA-1.5-7B, Qwen2.5-VL-7B, and InternVL3.5-8B use the same
input and structured output. These models are benchmark probes, not proposed
architectures. The evaluation splits are defined in Table~\ref{tab:splits}.}
\label{fig:overview}
\end{figure*}

\section{Introduction}
\label{sec:intro}

Vision models can recognize objects, scenes, and visual relations with high accuracy, but judging physical consistency requires a different form of understanding.
An image may appear realistic while containing evidence that conflicts with basic scene constraints. 
A cast shadow may disagree with the apparent illumination, a reflective surface may omit an object that should be visible, or an object may appear unsupported by the surface beneath it. 
Such failures are relevant to generated and edited imagery, but they also provide a useful test of whether visual models reason correctly about physical relations rather than surface appearance alone.

Evaluating this capability is difficult because physical inconsistencies are often sparse, local, and hard to annotate. Real images rarely provide ground-truth labels for the physical rule that is violated or the image region that provides the relevant evidence. 
Moreover, a high score on a synthetic test set does not necessarily show that a model has learned the intended physical relation. 
If training and test images are produced by the same rendering pipeline and intervention rules, the model may instead rely on visual artifacts associated with those procedures. 
This concern is especially important for generated benchmarks.

We study this problem as grounded physical consistency. Given a static RGB
image, the task is to determine whether its visible evidence is consistent with
a set of image-observable physical constraints; if it is not, the task is to
identify the inconsistency and localize the affected region.
We focus on constraints involving illumination and shadows, reflection, support and contact, surface response, and depth ordering. Rather than treating physical consistency as a single binary judgment, the benchmark records the physical constraint, the affected region, and the scene parameters used to construct each example.

To obtain controlled supervision, we generate paired physical counterfactuals from rendered scenes. 
Each physically consistent scene is paired with variants in which a targeted physical relation is changed while the remaining scene configuration is held fixed. 
This design provides category labels, bounding boxes, masks, and physical metadata without requiring manual annotation. 
The paired construction also lets us ask a harder question than
in-distribution classification: does a model recognize the physical constraint,
or does it learn shortcuts tied to the renderer and intervention procedure?

Our evaluation therefore covers both recognition and generalization.
In addition to held-out scene instances, we consider changes in scene appearance and violation-generation mechanisms, which provide stronger tests of whether the learned cues transfer beyond the conditions used to construct the training data. 
We used a filtered subset of PICA-100K for adaptation and reserve disjoint
PICABench samples for evaluation on physics-aware image editing.
PICABench evaluates physical realism across optical, mechanical, and
state-transition effects and provides region-grounded annotations and
questions, making it more closely aligned with physical-consistency evaluation
than a general image-editing dataset.

This work is an evaluation and benchmark study, not a new model or training method.
We use vision and vision-language models only as diagnostic
baselines. Their role is to measure benchmark difficulty and show how
performance changes across categories and controlled distribution shifts.

Our contributions are:

\begin{enumerate}
    \item We introduce a benchmark for grounded physical consistency in static images, with controlled violations including illumination and shadow, reflection, support and contact, surface response, and depth ordering.
    \item We provide renderer-derived supervision for affected regions, covering both introduced visual cues and expected locations of missing cues.
    \item We define evaluation settings that separate held-out scene recognition from appearance and intervention shifts, allowing us to test whether models learn physical constraints or rendering shortcuts.
    \item We provide a fixed protocol for 500 scene families and 5,500 images,
    and diagnostic baselines based on CLIP, LLaVA-1.5, Qwen2.5-VL, and
    InternVL3.5. Model adaptation is used only to probe the benchmark.
\end{enumerate}

Figure~\ref{fig:overview} summarizes benchmark construction and the shared
eval interface used by the evaluated VLMs.

\section{Related Work}
\label{sec:related}

\paragraph{Physical reasoning in vision and vision-language models.}
Recent benchmarks have examined whether vision-language models understand
physical properties, object relations, and scene dynamics. PhysBench evaluates
physical-world understanding across object properties, spatial relations,
scene-level reasoning, and dynamic events
\cite{PhysBench_ICLR2025}. Other diagnostic benchmarks have shown that strong
multimodal models can still fail on visually grounded reasoning and related
forms of multimodal understanding
\cite{hallusionbench_guan2024cvpr,NaturalBench_li2025,MMMU_Yue_2024_CVPR}.
These works motivate dedicated evaluation of physical understanding, but they
generally cover broad reasoning tasks rather than controlled image-level
physical inconsistencies with intervention-associated spatial supervision.
SPR-128K focuses more
specifically on spatial plausibility reasoning with multimodal language models
\cite{spr128k_hu2026}, while PhyDetEx studies the detection and explanation of
physical implausibility in generated videos
\cite{phydetex_wang2025}. Our focus is complementary: we study static images
through paired physical counterfactuals and evaluate whether models can identify
and spatially ground the violated physical constraint.

\paragraph{Physical consistency and image editing evaluation.}
Physical realism has become an important consideration in image generation and
editing. Prior work has shown that generated images can violate projective
geometry, object-shadow relations, and other physical constraints even when
their overall appearance is realistic
\cite{Shadows_dont_Lie_Sarkar_2024_CVPR}. 
Physics-aware editing benchmarks extend this observation by evaluating whether an edit preserves physical effects associated with the modified content.

PICABench is particularly relevant to this study. It evaluates physically realistic image editing through region-grounded questions covering optical, mechanical, and state-transition effects \cite{picabench_pu2026}.
Re-Edit considers image editing through a reasoning-aware evaluation framework \cite{reedit_ding2026},
while PhyEditBench studies physics-aware editing using real-world examples and a structured physical taxonomy \cite{phyeditbench_guo2026}. 

These works show that instruction following and perceptual quality alone are insufficient for evaluating edited images.
Our benchmark addresses this gap by isolating physical constraints through paired counterfactual rendering, providing exact intervention
metadata and spatial supervision, and supports controlled tests of whether model performance transfers beyond the rendering procedure used to construct the
data. We therefore select PICABench as the external evaluation source for
physics-aware edited images.

\paragraph{Manipulation detection and grounded analysis.}
Forgery detectors localize image changes using visual artifacts, physical
cues, or multimodal explanations
\cite{objectformer_wang2022cvpr,TruFor_Guillaro_2023_CVPR,
digital_forgeries,berkley_Photo_Manipulation,fakeshield_xu2025}.
Our task differs from authenticity detection: it tests a named physical
constraint and grounds the region affected by a controlled intervention.
Additional related work is summarized in Appendix~\ref{app:related}.

\paragraph{Synthetic counterfactual supervision and benchmark design.}
Synthetic rendering controls geometry, illumination, materials, and visibility,
but may introduce renderer- or intervention-specific shortcuts. We retain
paired counterfactuals for exact labels and spatial supervision, then separate
appearance shifts from changes in the intervention mechanism. The latter is
the stricter test of whether a model recognizes the intended relation.

\section{Task Definition}
\label{sec:task}

We focus on grounded physical consistency in a static RGB image $x$.
The task is to determine whether the visible scene evidence is consistent with
a set of image-observable physical constraints, identify the violated
constraint when an inconsistency is present, and localize its affected region.
We write the model output as
\begin{equation}
    f_\theta(x) = (\hat{y}, \hat{c}, \hat{r}, \hat{e}),
\end{equation}
where $\hat{y}\in\{0,1\}$ denotes physical consistency, $\hat{c}$ denotes the
predicted physical constraint, $\hat{r}$ is a normalized bounding box for the
affected region, and $\hat{e}$ is a natural-language text explanation of the
inconsistency. Physically
consistent images receive the clean label and no violation category.

The benchmark covers constraints related to illumination and shadows,
reflection, support and contact, surface response, and depth ordering. These
constraints are selected because their relevant evidence can be represented
visually under the controlled scene conditions used in the benchmark. We do
not treat the task as determining whether an image is physically possible
under every unobserved scene configuration. Instead, the goal is to evaluate
whether the visible cues are mutually consistent with the physical assumptions
defined for each benchmark group.

The ten labels are fine-grained sub-types within five physical families. Some
pairs are intentionally close, including wrong shadow direction versus
shadow illumination inconsistency and floating object versus impossible
occlusion. Their operational boundaries and shortcut risks are listed in
Appendix~\ref{tab:taxonomy}; we do not assume that the labels are independent.

The spatial target localizes the region affected by the intervention. For an
introduced cue, such as a displaced shadow, it covers the anomalous region; for
a missing cue, it covers the region where that cue should appear. This target
need not contain every contextual cue required to diagnose the violation. We
therefore distinguish affected-region localization from the stricter problem
of complete evidence grounding in Section~\ref{sec:benchmark_dataset}.

\section{Counterfactual Physical Consistency Benchmark}
\label{sec:benchmark_dataset}

\paragraph{Controlled scene construction.}
Our evaluation uses multiple image sources; here we describe the
Mitsuba~3 pipeline used to generate the controlled subset. It contains 500 scene families and 5,500
images: one clean correct image and ten counterfactual images per family. Mitsuba~3
\cite{Mitsuba3} scenes are deterministic from a recorded seed. Families or categories vary
the room, object geometry, pose, material, illumination, and camera. We render
$500\times500$ images; the full set takes about ten hours on one NVIDIA A100
GPU with ten CPU cores and 50~GB of system memory. Renderer and sampling
details appear in Appendix~\ref{app:implementation}.

\begin{table}[t]
\centering
\small
\setlength{\tabcolsep}{4pt}
\begin{tabular}{@{}lrrl@{}}
\toprule
Partition & Scenes & Images & Role \\
\midrule
Training & 300 & 3,300 & adaptation \\
Validation & 50 & 550 & model selection \\
Standard held-out & 50 & 550 & unseen scene seeds \\
Appearance shift & 50 & 550 & unseen visual factors \\
Intervention shift & 50 & 550 & unseen construction rules \\
\midrule
Total & 500 & 5,500 & \\
\bottomrule
\end{tabular}
\caption{Scene-disjoint synthetic benchmark partitions. Every scene family
contains one image for each of the 11 labels.}
\label{tab:splits}
\end{table}

Each label therefore has 300 training, 50 validation, and 150 test images,
with 50 test images in each suite. Models may use only the training split for
adaptation and the validation split for selection. All three test suites use a
fixed prompt, parser, and checkpoint. The intervention-shift macro-F1 is the
primary result for the question posed in the title.

\paragraph{Matched physical counterfactuals.}
The subset covers wrong shadow direction, missing cast shadow, floating
objects, shadow illumination inconsistency, missing reflection,
reflection angle mismatch, incorrect specular highlight, wrong reflected
object identity, impossible occlusion order, and missing contact shadow.
Violations are produced either by intervening on scene parameters or by
compositing auxiliary renders in linear floating-point RGB radiance. For
example, we lift an object, change illumination, transfer a shadow rendered
under an alternative light position, or replace content visible through a
reflective surface. Scene parameters unrelated to the targeted relation are
preserved as closely as possible.

Some categories require different scene preconditions. For example, reflection violations
use a reflective back wall, whereas impossible occlusion introduces two objects
at different depths. We therefore render a category-specific clean reference
to validate each counterfactual. The scored set contains one
clean item and ten violated items per family. Each pair shares the same seed
and nuisance variables within its category subscene.

\paragraph{Appearance and intervention shifts.}
The appearance shift partition uses object geometries, material configurations,
and lighting combinations that are disjoint from training while retaining the
same category definitions and counterfactual mechanisms. The
intervention-shift partition instead matches the training appearance
distribution but holds out the construction mechanism for each semantic
violation. For example, a shadow-direction violation can be generated by
changing illumination geometry or by transferring an auxiliary-render shadow;
only one mechanism is observed during training. Across both shifts, scene
families remain disjoint, categories are balanced, and clean/counterfactual
pairs share the same nuisance variables. We additionally match affected-region
area and counterfactual effect size across mechanisms to reduce low-level
artifact cues.

Composited counterfactuals can still contain boundary halos, small color
changes, or other editing cues. We therefore treat intervention shift as a
required result rather than an optional robustness test. We also audit
difference images and report performance by construction mechanism. These
checks do not prove that every shortcut has been removed; they make the most
obvious shortcuts visible in the evaluation.

\paragraph{Renderer-derived annotations.}
For each pair, the pipeline derives its clean/violated status, category,
procedural scene parameters, intervention metadata, a templated physical
explanation, an affected-region mask, and its tight axis-aligned bounding box.
A semantic
visibility integrator marks pixels whose primary ray intersects named geometry,
and a cast-shadow integrator marks visible receiver points for which the target
object blocks the point light. Other categories use render differences or
known compositing regions. These geometry-derived passes avoid extracting
spatial targets by thresholding the final RGB image.

We add an intervention mask containing pixels directly changed
by construction, from an evidence mask, containing all visual evidence
needed for diagnosis. The stored \texttt{affected\_mask} represents the former;
the two coincide for some categories but not necessarily for cases such as a
floating object, where the air gap is diagnostically important. We therefore
describe current spatial supervision as affected-region localization and do
not assume that it is a complete evidence annotation.

\paragraph{Validation.}
Generation validates object and cast-shadow visibility, mask and bounding-box
area, mask fragmentation, and large boundary changes around composited
regions. Sampling is deterministic with respect to the
recorded scene seed, and we group all samples from the same underlying scene
family when constructing data partitions. Counterfactual construction occurs
in linear radiance before image conversion. In the final release, aligned raw
images and masks are retained, while independently denoised outputs are
reserved for presentation. Output metadata will record the Mitsuba and Dr.Jit
versions, renderer variant, hardware backend, render seed, integrator and
sampling configuration, image-processing settings, and generator version.

\paragraph{PICA-20K adaptation subset.}
We select 20,000 valid, visibly grounded examples from PICA-100K. Source videos
and edit identities are disjoint between adaptation and evaluation. Selection
fields and reconstruction details appear in Appendix~\ref{app:implementation}.

Additional clean/counterfactual pairs for every category appear in
Appendix~\ref{app:examples}. They show variation across objects and camera
settings and make residual rendering or compositing cues easier to audit.

\section{Benchmark Models and Adaptation}
\label{sec:models}

We evaluate vision and vision-language models as benchmark baselines rather
than proposing a new model architecture. The probe set includes
LLaVA-1.5-7B \cite{llava_v1}, Qwen2.5-VL-7B-Instruct
\cite{qwen25vl_bai2025}, and InternVL3.5-8B \cite{internvl35_wang2025}.
These models provide different visual encoders, image-resolution policies, and
language backbones at a similar scale. We evaluate the pretrained models and
adapt each one with LoRA \cite{lora_hu2022iclr}. The adaptation is a common
measurement tool; it is not a proposed learning method.

Each training example contains an image and an instruction requesting a
structured physical-consistency diagnosis. The target specifies consistency,
the physical constraint when applicable, the affected region, and an
explanation. Prompt tokens are excluded from the training loss so that
optimization is applied to the diagnostic response.

The spatial output is a normalized $(x_{\min},y_{\min},x_{\max},y_{\max})$
box. Renderer masks define and validate the target box but are not model
outputs. Coordinate conversion and validation are given in
Appendix~\ref{app:implementation}.

We train a synthetic-only model for zero-shot PICABench transfer and a
mixed-source model that additionally uses PICA-20K. The latter converts each
grounded question--answer pair and its box into the same structured response
format. The two variants are evaluated separately to distinguish transfer from
adaptation.

All three models use the same data, response schema, validation rule, and
three-seed protocol. Hyperparameters, preprocessing, and adapter targets are
listed in Appendix~\ref{app:implementation}.

\section{Experiments}
\label{sec:experiments}

\paragraph{Setup.}
We separate four evaluation settings. The standard held-out split measures
in-distribution generalization to unseen scene identities. The
appearance shift holds out object geometries, materials, and lighting
configurations while preserving the intervention mechanisms. The
intervention shift renders the same semantic violations using mechanisms
not observed during training while matching the scene-appearance distribution.
Finally, PICABench evaluation uses its native region-grounded questions rather
than forcing its broader physical taxonomy into our ten synthetic categories.
We report both zero-shot transfer from synthetic-only training and adaptation
with PICA-20K on a disjoint held-out subset. Model interpretation and causal
analysis are outside the scope of this benchmark study.

The combined synthetic test contains 1,650 images from 150 unseen scene
families: 550 images in each of the standard, appearance-shift, and
intervention-shift suites. The model generates structured text, from which we parse the
binary decision, one of ten violation categories or clean, the affected region,
and the explanation. We report binary accuracy, binary balanced accuracy,
binary macro-F1, both binary class recalls, 11-class accuracy and macro-F1,
box IoU, invalid-output rate, and explanation correctness. Invalid or missing
spatial outputs receive zero localization score and also count as invalid.
Adapted results average three random seeds. A human reference study uses three
raters on the same stratified image set without renderer metadata.

\paragraph{Baselines.}
The comparison includes three reference baselines. The majority class baseline always
predicts the most frequent class and provides a trivial lower-bound reference
under class imbalance. CLIP ViT-L/14 zero-shot matches each
image against text descriptions of all eleven categories using cosine
similarity of CLIP image and text embeddings, without any task-specific
training. LLaVA-1.5-7B zero-shot applies the base model (no LoRA
adapters) with the same structured prompt and parser.
The three LoRA-adapted VLMs are diagnostic probes trained with the common
protocol in Section~\ref{sec:models}.

\paragraph{Standard held-out results.}

\begin{table*}[t]
\centering
\small
\setlength{\tabcolsep}{3.5pt}
\begin{tabular}{@{}lrrrrrrr@{}}
\toprule
Method & Bin.~Acc & Bal.~Acc & Bin.~mF1 & Clean R & Viol.~R & Cat.~Acc & Cat.~mF1 \\
\midrule
Majority class          & 90.9 & 50.0 & 47.6 & 0.0 & 100.0 & 9.1 & 1.5 \\
CLIP zero-shot          & 13.2 & 27.7 & 13.0 & 45.3 & 10.0 & 6.2 & 1.9 \\
LLaVA-1.5-7B zero-shot  & 88.4 & 49.5 & 48.4 & 2.0 & 97.0 & 1.2 & 1.2 \\
\midrule
LLaVA-1.5-7B + LoRA     & 98.4 & 92.7 & 94.8 & 85.8 & 99.6 & 63.7 & 64.0 \\
Qwen2.5-VL-7B + LoRA    & 98.6 & 92.5 & 95.3 & 85.1 & 99.9 & 68.1 & 67.8 \\
InternVL3.5-8B + LoRA   & 97.9 & 91.2 & 93.3 & 83.0 & 99.4 & 66.3 & 66.0 \\
\bottomrule
\end{tabular}
\caption{Diagnostic baselines on the 550-image standard held-out split. All
scores are percentages; mF1 denotes macro-F1. Adapted-model entries are means
over three seeds.}
\label{tab:baselines}
\end{table*}

Zero-shot CLIP and LLaVA perform poorly on the 11-class task. LLaVA also
over-predicts violations, as shown by its 2.0\% clean recall. Adaptation makes
the labels learnable for all three VLMs, with Qwen2.5-VL giving the strongest
category macro-F1 (67.8\%). This result alone does not show physical reasoning;
that claim depends on the shift results in Table~\ref{tab:generalization}.

The 90.9\% majority accuracy follows directly from the ten violated images for
every clean image. It is therefore not evidence of useful detection. Balanced
accuracy, binary macro-F1, and the separate clean and violated recalls are the
primary binary measures; raw binary accuracy is retained for completeness.

\paragraph{Per-category results.}
Appendix~\ref{app:pilot} gives a category breakdown and confusion matrix for
the earlier 1,650-image development pilot. It exposes complete confusion within
two deliberately close pairs: shadow direction versus illumination consistency,
and impossible occlusion versus floating. These results motivate reporting
both broad detection and fine-grained categories.

\paragraph{Generalization beyond held-out scenes.}
Table~\ref{tab:generalization} organizes the measurements needed to distinguish
constraint learning from procedural shortcuts. LLaVA loses 11.6 category-F1
points under appearance shift and 23.2 points under intervention shift. The
larger intervention gap indicates substantial sensitivity to how a violation
is constructed. PICABench further separates synthetic transfer from adaptation
to real edited images.

\begin{table*}[t]
\centering
\small
\setlength{\tabcolsep}{3pt}
\begin{tabular}{@{}lccccc@{}}
\toprule
Setting & Bin.~mF1 & Cat.~mF1 & Box IoU (\%) &
Invalid (\%) & Expl./QA correct (\%) \\
\midrule
Standard held-out             & 94.8 & 64.0 & 60.5 & 2.7 & 77.4 \\
Appearance shift              & 86.7 & 52.4 & 50.6 & 5.1 & 68.3 \\
Intervention shift            & 77.9 & 40.8 & 41.7 & 7.6 & 58.9 \\
PICABench zero-shot           & n/a & n/a & 22.7 & 14.2 & 38.6 \\
PICABench adapted (+PICA-20K) & n/a & n/a & 34.8 & 8.6 & 49.1 \\
Human reference (standard)    & 96.0 & 82.0 & 74.0 & n/a & 91.0 \\
\bottomrule
\end{tabular}
\caption{LLaVA-1.5-7B + LoRA across evaluation settings, with human performance
as a reference. Scores are percentages and average three seeds. \emph{n/a}
denotes a metric incompatible with the native PICABench task.}
\label{tab:generalization}
\end{table*}

\paragraph{Affected-region and explanation evaluation.}
We compute box IoU between the normalized predicted box and the tight box
derived from the renderer's \texttt{affected\_mask}. This measures localization
of the intervention-associated region, not complete causal evidence grounding.
For PICABench, we normalize its $(x,y,w,h)$ question boxes using the edited
image dimensions and evaluate against the box associated with each native
question. Malformed responses, missing fields, non-positive or out-of-range
coordinates, and unrecognized categories contribute to the invalid-output
rate; failed spatial predictions receive zero IoU.

Explanation correctness is assessed independently of wording using a
three-level rubric: incorrect, partially correct, or correct identification of
the violated relation and its visible consequence. We report the fraction
that receive a majority ``correct'' rating from three raters. Agreement is
substantial (Fleiss' $\kappa=0.78$). Human performance uses a category-balanced
subset of 220 images. A fixed qualitative panel contains
successful, ambiguous, and failed box-grounded predictions from the synthetic
shifts and PICABench.

We compute 95\% bootstrap confidence intervals by
resampling scene families, not individual images. This preserves dependence
between each clean image and its matched counterfactuals. Across synthetic
suites, interval half-widths are at most 1.8 points for binary macro-F1, 2.6
for category macro-F1, and 2.4 for box IoU. Across three seeds, the respective
standard deviations are at most 0.6, 0.9, and 1.1 points.

\section{Discussion, Limitations, and Ethics}
\label{sec:discussion}

The standard and shift results show that a VLM can fit several rendered
violation labels, but does not retain this accuracy under controlled changes.
Categories with direct visual evidence are easier than closely related shadow
and geometry sub-types. This is evidence of benchmark learnability, not of
general physical reasoning.

The CLIP zero-shot result is instructive: text-image similarity alone performs
poorly on the standard split (1.9\% category macro-F1). Adaptation is necessary,
but the 23.2-point intervention gap shows that it still learns cues tied to
counterfactual construction. PICA-20K adaptation improves external box IoU by
12.1 points and QA correctness by 10.5 points, but both remain well below the
synthetic standard split.

Affected-region masks provide a reproducible localization target but do not
always capture every cue needed for causal diagnosis. We therefore report them
as intervention-associated localization and evaluate explanations separately.
PICABench complements this controlled target with region-grounded questions on
edited images, although its broader taxonomy is not directly interchangeable
with our ten synthetic categories.

The synthetic construction also leaves visible risks. In particular, removed
contact shadows can create boundary or halo cues, and some auxiliary-render
composites may change more than the intended relation. The paired design and
area matching reduce these cues but cannot remove them completely. For this
reason, the intervention-shift result and mechanism-level breakdowns are
central to the main claim.

A high violation score should not be treated as proof that an image was
forged, generated, or edited with malicious intent. We position the system as
a diagnostic tool for editing pipelines and dataset auditing, not as an
authenticity oracle.

\section{Conclusion}
\label{sec:conclusion}

We introduced a counterfactual benchmark for grounded physical consistency.
Its 500 scene families provide 5,500 clean and violated images with category,
region, metadata, and language supervision. Adapted VLMs learn the standard
split, but LLaVA category macro-F1 drops from 64.0\% to 40.8\% when the
intervention mechanism changes. The answer to the title question is therefore
mixed: the model learns useful physical categories, but also relies strongly on
construction-specific cues. External adaptation helps, yet a large transfer
gap remains. Future work will study internal model representations, causal
interventions on model features, and the faithfulness of generated
explanations. Those questions are outside this study.

{\small
\bibliographystyle{plainnat}
\bibliography{refrences}
}

\appendix

\section{Additional Related Work}
\label{app:related}

Image manipulation detectors use high-frequency features, object consistency,
and diffusion priors to detect and localize changes
\cite{objectformer_wang2022cvpr,TruFor_Guillaro_2023_CVPR,
UnionFormer_Li_2024_CVPR,DiffForensics_Yu_2024_CVPR}. Physical approaches
instead test illumination or shadow consistency
\cite{digital_forgeries,berkley_Photo_Manipulation}. FakeShield, ForgeryGPT,
and ForenX add grounded language explanations
\cite{fakeshield_xu2025,ForgeryGPT_zhang2026,ForenX_tan2025}. These methods
address authenticity or manipulation. Our benchmark instead labels the
physical relation changed by a controlled intervention.

\section{Dataset Examples and Category Boundaries}
\label{app:examples}

\begin{figure*}[t]
\centering
\includegraphics[width=\textwidth]{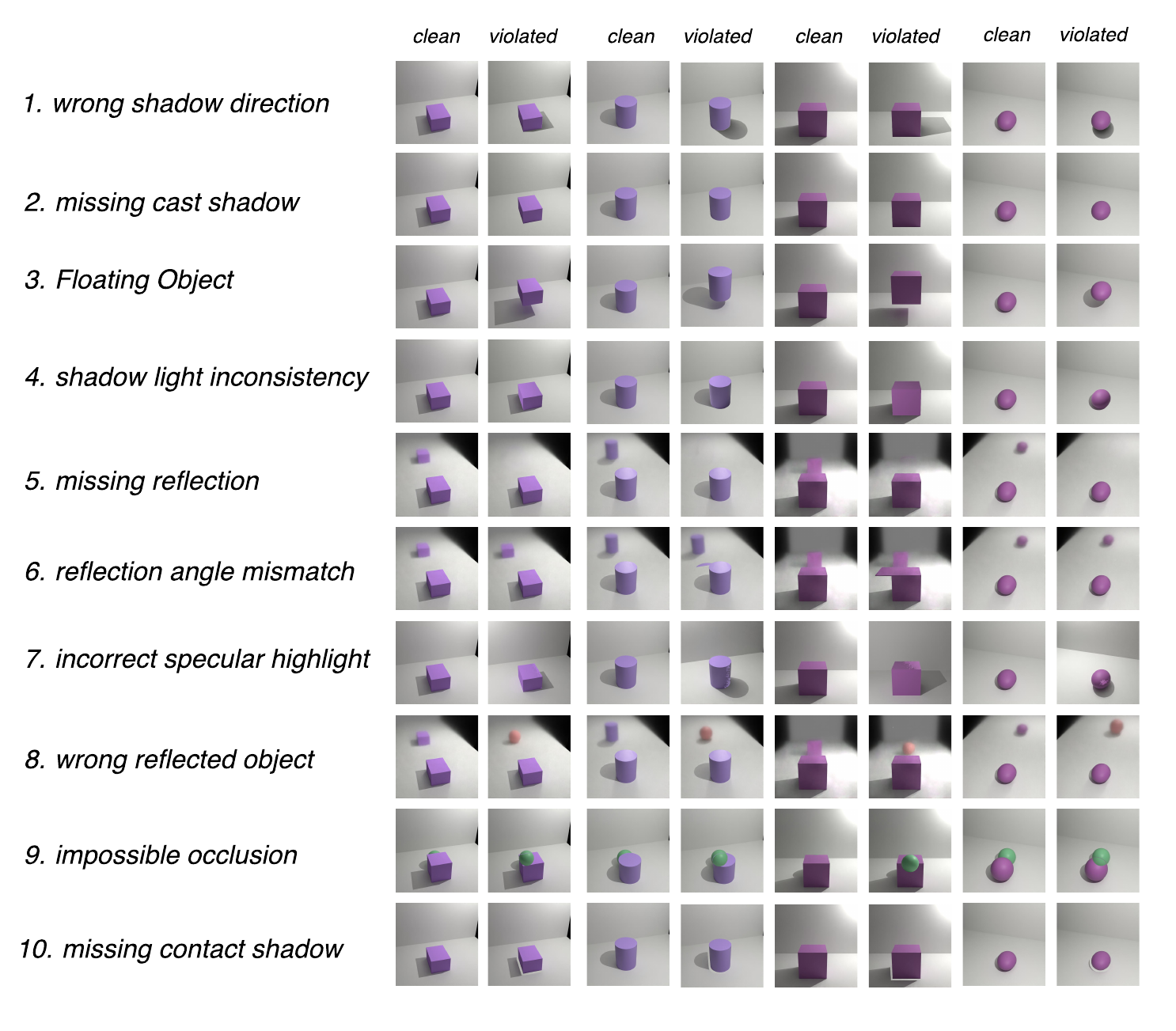}
\caption{Additional clean and violated examples for all ten fine-grained
categories. Columns vary object geometry, camera view, lighting, and scene
appearance. The panel is also used for visual auditing: cues such as hard
boundaries, halos, or unrelated color changes are recorded rather than hidden.}
\label{fig:appendix_examples}
\end{figure*}

\begin{table*}[t]
\centering
\small
\setlength{\tabcolsep}{4pt}
\begin{tabular}{@{}p{0.21\textwidth}p{0.44\textwidth}p{0.27\textwidth}@{}}
\toprule
Fine label & Operational distinction & Main shortcut risk \\
\midrule
Wrong shadow direction & Shadow direction changes while object shading and scene lighting are fixed. & Transferred-shadow boundary \\
Missing cast shadow & The object and contact cue remain, but its cast shadow is removed. & Local smoothing or removal trace \\
Floating object & A visible gap separates the object from its support; other lighting remains valid. & Gap or silhouette alone \\
Shadow--illumination inconsistency & Object shading comes from one light setting and the cast shadow from another. & Close to wrong shadow direction \\
Missing reflection & A visible object that should appear in a verified reflective region is absent. & Reflection-region texture change \\
Reflection angle mismatch & Reflected content is displaced from its mirror-consistent position. & Composite edge or displacement \\
Incorrect specular highlight & Highlight position conflicts with fixed light, view, and surface geometry. & Broad material or color change \\
Wrong reflected object & Reflected identity differs from the visible source object. & Object identity alone \\
Impossible occlusion & Local visibility reverses the recorded front-to-back depth order; support is unchanged. & Interpenetration boundary \\
Missing contact shadow & The object stays in contact and the larger cast shadow remains, but the near-contact cue is removed. & Bright rim or halo \\
\bottomrule
\end{tabular}
\caption{Operational category definitions. The close shadow and geometry
pairs are intentional stress tests. Human agreement and family-level metrics
will be reported so that these sub-types are not treated as independent facts.}
\label{tab:taxonomy}
\end{table*}

\section{Baseline and Evaluation Details}
\label{app:implementation}

Scenes use a ground plane, back wall, ceiling, and side walls. We sample
spheres, cubes, non-uniform boxes, capped cylinders, and ellipsoids, together
with pose, scale, color, roughness, and diffuse or GGX rough-plastic material.
Reflectors use a low-roughness conductor. Illumination combines a point emitter
with a weak constant environment; light elevation spans $30^\circ$--$70^\circ$
and azimuth uses eight directions. The perspective camera has a $38^\circ$
horizontal field of view. Images use 150 samples per pixel, path depth 8, and
a box reconstruction filter.

LLaVA-1.5-7B uses its CLIP ViT-L/14-336 visual encoder and Vicuna-7B language
backbone. Qwen2.5-VL-7B-Instruct and InternVL3.5-8B use their official image
processors and native dynamic-resolution settings. For a controlled probe, the
visual encoder is frozen and LoRA is applied to language attention projections;
the response schema, train/validation split, stopping rule, and parser are
shared. The release records the exact checkpoint revision, processor version,
adapter targets, random seed, and maximum visual-token count for each model.

For LLaVA, LoRA rank is 32, alpha is 64, and dropout is 0.05. We train for
three epochs with AdamW, learning rate $2\times10^{-4}$, cosine annealing,
batch size 4, gradient accumulation 2, maximum sequence length 512, and
bfloat16 precision on one A100. The lowest validation loss selects the
checkpoint. PICABench pixel boxes $(x,y,w,h)$ are converted to
$[x/W,y/H,(x+w)/W,(y+h)/H]$ and checked for positive extent and image bounds.

For PICA-20K, each record retains the input and edited-image paths, instruction,
physics category and law, operation, difficulty, grounded question--answer
pairs, and edit-area boxes. The released selection file reports category counts
and source identities so the filtered subset can be reconstructed without
leakage.

The human reference subset contains 220 images, with 20 examples per label and
no repeated scene family across the sampling units. Three raters receive the
same task definition as the models but no renderer metadata. They provide a
binary decision, a fine category, one affected-region box, and a short reason.
We report majority-vote performance, Fleiss' $\kappa$ for categorical labels,
and agreement on boxes at IoU thresholds 0.5 and 0.75.

\section{Development-Pilot Category Analysis}
\label{app:pilot}

Before locking the three test suites, we evaluated LLaVA-1.5-7B + LoRA on an
earlier 1,650-image in-distribution development pool. This pool is not used for
the headline comparison. It is retained here because it reveals the intended
fine-category stress cases.

\begin{table}[t]
\centering
\small
\setlength{\tabcolsep}{5pt}
\begin{tabular}{@{}lcc@{}}
\toprule
Category & Recall & F1 \\
\midrule
Floating object              & 0.993 & 0.997 \\
Incorrect specular highlight & 0.993 & 0.997 \\
Missing contact shadow       & 0.980 & 0.990 \\
Wrong reflected object       & 0.993 & 0.987 \\
Clean                        & 0.867 & 0.919 \\
Missing cast shadow          & 1.000 & 0.804 \\
Reflection angle mismatch    & 0.800 & 0.723 \\
Missing reflection           & 0.393 & 0.539 \\
Wrong shadow direction       & 0.120 & 0.214 \\
Shadow--illumination inconsistency & 0.000 & 0.000 \\
Impossible occlusion         & 0.000 & 0.000 \\
\midrule
\textbf{Macro average}       & --    & \textbf{0.652} \\
\bottomrule
\end{tabular}
\caption{Per-category recall and F1 on the 1,650-image development pilot.}
\label{tab:percategory}
\end{table}

\begin{figure}[t]
\centering
\includegraphics[width=0.88\textwidth]{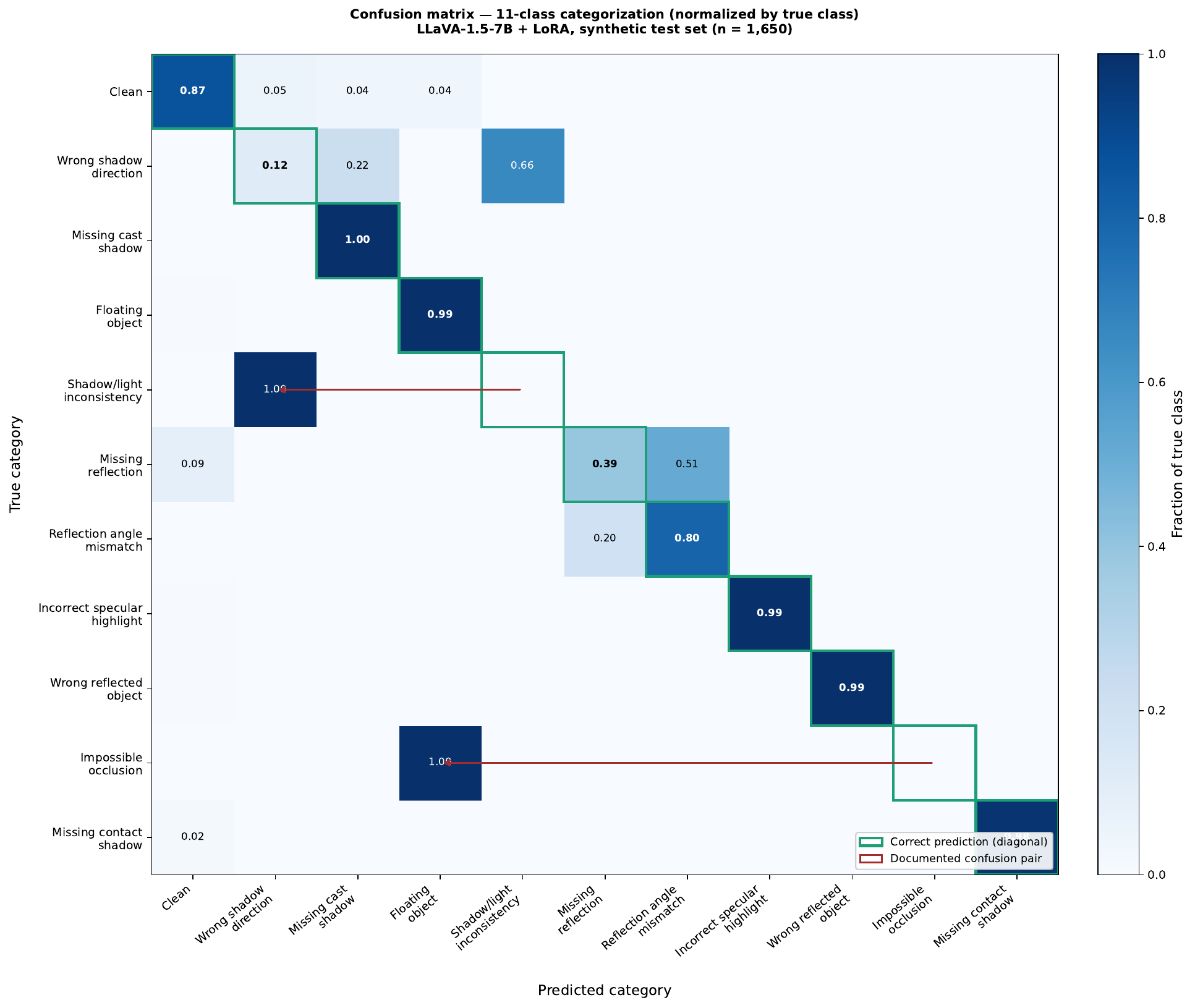}
\caption{Development-pilot confusion matrix. Shadow--illumination
inconsistency is absorbed by wrong shadow direction, while impossible
occlusion is absorbed by floating object. These close pairs motivate the
operational definitions in Table~\ref{tab:taxonomy}.}
\label{fig:confusion}
\end{figure}

\end{document}